\documentclass{article}

\usepackage[preprint]{neurips_2026}

\usepackage[utf8]{inputenc}
\usepackage[T1]{fontenc}
\usepackage{hyperref}
\hypersetup{
  colorlinks=false,
  pdfborder={0 0 1},
  citebordercolor={1 0 0},
  linkbordercolor={1 0 0},
  urlbordercolor={1 0 0},
}
\usepackage{url}
\usepackage{booktabs}
\usepackage{amsfonts}
\usepackage{amsmath}
\usepackage{amssymb}
\usepackage{pifont}
\usepackage{nicefrac}
\usepackage{microtype}
\usepackage{xcolor}
\usepackage{graphicx}
\usepackage{subcaption}
\usepackage{algorithm}
\usepackage{algpseudocode}
\usepackage{multirow}
\usepackage{tikz}
\usetikzlibrary{shapes.geometric,shapes.symbols,arrows.meta,positioning,fit,backgrounds,calc,matrix}
\usepackage{pgfplots}
\usepackage{enumitem}
\pgfplotsset{compat=1.9} 

\title{Latent Clarity: Bridging World-Model Kinematics to Semantic Manifolds\\
for Video Anomaly Anticipation}

\author{%
  Abu Anas Ibn Samad \\
  Sigmind.ai  \\
  Dhaka 1215, Bangladesh \\
  \texttt{abushuvom@sigmind.ai}, \texttt{abushuvom@gmail.com} \\
}

\begin{document}

\maketitle

\begin{abstract}
Continuous video anomaly detection is dominated by reactive Multiple Instance
Learning (MIL) formulations that collapse dense spatiotemporal features into
scalar anomaly scores. We introduce \textbf{PULS} (Predictive Unified Latent
Space), a continuous semantic world-model pipeline comprising two core modules: 
a 490M-parameter \textbf{KSD Bridge} (Kinematic-to-Semantic Distillation) and 
a 16.8M-parameter \textbf{Anticipatory State Predictor (ASP)}. The KSD Bridge 
translates V-JEPA\,2 physical tensors into the 2048-d \textsc{Qwen3-VL-Embedding-2B} 
text-aligned hypersphere via symmetric InfoNCE with SALT noise injection, trained 
on a subset of \emph{UCF-Crime} dataset. This semantic translation alone yields 
a chunk-level AUROC of $\mathbf{0.8994}$ for UCF-Crime and $\mathbf{0.8162}$ for out-of-distribution XD-Violence validation set without MIL or hierarchical fusion.

Crucially, we introduce and empirically validate the \emph{Latent Clarity Hypothesis}:
because JEPA's temporal predictor discards aleatoric pixel noise (motion blur,
occlusion) while preserving continuous kinematics, hallucinated future
representations are inherently more semantically separable than observed
presents. The ASP, trained via symmetric InfoNCE to sharpen these hallucinated 
future latents within the hypersphere, achieves $\mathbf{44.5\%}$ mean 14-way 
zero-shot VQA accuracy, exceeding the observation baseline ($34.9\%$) by $+9.6$\,pp. 
Applying the ASP to Observation Tensors collapses accuracy to $7.3\%$ (random chance), 
proving that Anticipation and Observation occupy geometrically distinct sub-manifolds.
A \textbf{Triple-Track Lead-Time} protocol with an $L_1$-surprise gate yields
a peak $+8.9$\,pp anticipatory advantage at $T{-}0.5$s ($p{<}0.001$,
$N{=}1{,}000$ permutation), structurally separating temporal-physics
anticipation from static scene priors. Zero-shot transfer to XD-Violence
confirms that Newtonian-invariant kinematic representations generalize
out-of-distribution for open vocabulary semantic reasoning capability.
\end{abstract}

\section{Introduction}
\label{sec:intro}

Video anomaly detection (VAD) in continuous surveillance streams is almost
universally cast as Multiple Instance Learning (MIL) over weakly-labeled
bags of frames~\cite{sultani2018,rtfm,s3r,dakd}. These architectures consume
dense optical or semantic features and regress a scalar anomaly score per
frame, enabling rank-based Area Under the ROC Curve (AUROC) evaluation
against the binary ground truth.
Although MIL scores have driven steady AUROC progress, the paradigm is
structurally \emph{reactive}: anomalies are detected only \emph{after} their
pixel-level manifestation, and the scalar output carries no semantics of
\emph{what} happened, only that \emph{something} happened. Replacing the
scalar with generative vision--language modeling does not solve the latency
problem: autoregressive token generation at every sliding window is
prohibitively slow for real-time edge deployment.

Joint-Embedding Predictive Architectures
(JEPAs)~\cite{lecun_path,ijepa,vjepa2,vljepa} offer a principled alternative.
Instead of reconstructing pixels or tokens, V-JEPA\,2 predicts the
\emph{latent} representations of masked future patches in an abstract
representation space, discarding unpredictable surface-level details. The
resulting features are temporally structured, compact, and carry intuitive
physics. With VLJEPA bridges V-JEPA\,2's continuous
latent physics into a text-aligned semantic manifold at inference speed
compatible with continuous video, however, it's still reactive for anomaly sensing with static scene bias (Section~\ref{sec:discuss}).

This paper proposes and empirically validates such a bridge under the
\textbf{PULS} (Predictive Unified Latent Space) framework.
Concretely, our contributions are:

\begin{itemize}
 \item \textbf{KSD Bridge.} A 490M-parameter bridge (\textsc{LLaMA}-8L, bidirectional) mapping the V-JEPA\,2 Observation Tensor $[\mathbf{Z}_{\mathrm{ctx}} \,;\, \mathbf{Z}_{\mathrm{target}}]$ into the \textsc{Qwen3-VL} Hypersphere ($\mathcal{S}^{2047}$) via SALT-augmented InfoNCE. Yields chunk-AUROC $\mathbf{0.8994}$ on UCF-Crime without MIL (Section~\ref{sec:ksd}).
  
  \item \textbf{Anticipatory State Predictor (ASP).} A 16.8M-parameter residual MLP resolving epistemic blur. It routes the Anticipation Tensor $[\mathbf{Z}_{\mathrm{ctx}} \,;\, \hat{\mathbf{Z}}_{\mathrm{fut}}]$ through the frozen KSD Bridge, sharpening it to outperform the observation baseline by $+9.6$pp VQA accuracy, validating the \emph{Latent Clarity Hypothesis} (Sections~\ref{sec:asp}, \ref{sec:vqa}).
  
  \item \textbf{Cross-Manifold Probe.} Passing Observation Tensors through the ASP collapses Q1 VQA accuracy to $7.3\%$ and binary AUROC to $0.5535$ --- proving Anticipation and Observation sub-manifolds are geometrically distinct (Section~\ref{sec:cmp}).
  
  \item \textbf{Triple-Track Lead-Time Protocol.} A framework to decompos scene bias from temporal physics via ungated, $L_1$-surprise-gated, and context-only tracks. Yields a peak $+8.9$pp anticipatory advantage at $T{-}0.5$s ($p{<}0.001, N{=}1000$); $0\%$ calm-zone saturation proves system forecasts true future kinematics rather than overfitting background priors (Section~\ref{sec:leadtime}).

  \item \textbf{Zero-Shot OOD Transfer.} Without retraining, PULS transfers to XD-Violence (dynamic cameras), retaining chunk-AUROC $0.816$ and a $+4.5$pp lead-time advantage at $T{-}1.5$s (Section~\ref{sec:xd}).
\end{itemize}

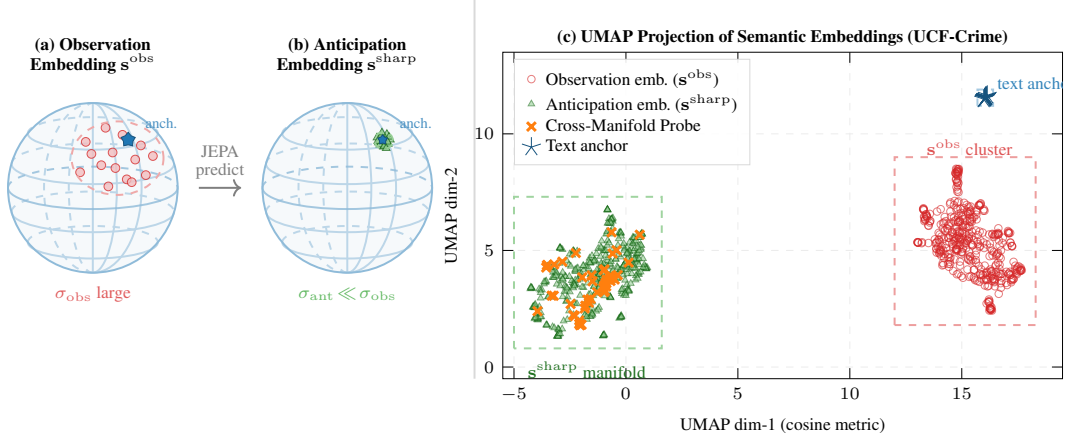
\begin{figure*}[t]
\centering

\definecolor{nblue}{RGB}{31,119,180}
\definecolor{nred}{RGB}{214,39,40}
\definecolor{ngreen}{RGB}{44,160,44}
\definecolor{norange}{RGB}{255,127,14}
\definecolor{ngray}{RGB}{127,127,127}

\resizebox{\textwidth}{!}{%
\begin{tikzpicture}[
  every node/.style={font=\scriptsize},
]


\begin{scope}[shift={(1.4, 3.5)}]
  \node[font=\bfseries\scriptsize, align=center, anchor=south] at (0,1.5)
    {(a) Observation \\ Embedding $\mathbf{s}^{\mathrm{obs}}$};

  \fill[nblue!5, opacity=0.8] (0,0) circle (1.2);
  
  \foreach \y/\xr in {0.34/1.15, 0.77/0.92, -0.34/1.15, -0.77/0.92} {
    \draw[nblue!30, thick] (\xr,\y) arc (0:-180:{\xr} and {\xr*0.25});
    \draw[nblue!30, thick, dashed] (-\xr,\y) arc (180:0:{\xr} and {\xr*0.25});
  }
  \foreach \x in {0.34, 0.69} {
    \draw[nblue!30, thick] (0,1.2) arc (90:-90:{\x} and 1.2);
    \draw[nblue!30, thick, dashed] (0,-1.2) arc (270:90:{\x} and 1.2);
  }
  \draw[nblue!30, thick] (0,1.2) -- (0,-1.2);
  
  \draw[nblue!50, thick, opacity=0.8] (1.2,0) arc (0:-180:1.2 and 0.3);
  \draw[nblue!50, thick, dashed, opacity=0.8] (-1.2,0) arc (180:0:1.2 and 0.3);

  \draw[nblue!50, thick] (0,0) circle (1.2);

  \foreach \x/\y in {
    0.18/0.85, 0.60/0.40, 0.28/0.50, 0.75/0.18, 0.42/0.75,
    0.12/0.32, 0.68/0.60, -0.10/0.65, 0.48/0.12, 0.85/0.45,
    0.32/0.28, -0.02/0.48, 0.55/0.08, -0.18/0.22, 0.22/0.02
  } {
    \node[draw=nred, fill=nred!30, circle, minimum size=3.5pt, inner sep=0pt, opacity=0.8] at (\x,\y) {};
  }

  \draw[nred!40, dashed, thick] (0.35,0.42) ellipse (0.65 and 0.52);

  \node[draw=nblue!70!black, fill=nblue, star, star points=5, minimum size=6pt,
        inner sep=0pt] at (0.50,0.68) {};
  \node[font=\tiny, nblue!80, anchor=south west] at (0.62,0.72) {anch.};

  \node[font=\scriptsize, nred!75] at (0,-1.5) {$\sigma_{\mathrm{obs}}$ large};
\end{scope}

\draw[->, thick, ngray] (2.9, 3.5) -- (3.5, 3.5)
  node[midway, above, font=\scriptsize, align=center] {JEPA\\[-1pt]predict};

\begin{scope}[shift={(5.0, 3.5)}]
  \node[font=\bfseries\scriptsize, align=center, anchor=south] at (0,1.5)
    {(b) Anticipation \\ Embedding $\mathbf{s}^{\mathrm{sharp}}$};

  \fill[nblue!5, opacity=0.8] (0,0) circle (1.2);
  
  \foreach \y/\xr in {0.34/1.15, 0.77/0.92, -0.34/1.15, -0.77/0.92} {
    \draw[nblue!30, thick] (\xr,\y) arc (0:-180:{\xr} and {\xr*0.25});
    \draw[nblue!30, thick, dashed] (-\xr,\y) arc (180:0:{\xr} and {\xr*0.25});
  }
  \foreach \x in {0.34, 0.69} {
    \draw[nblue!30, thick] (0,1.2) arc (90:-90:{\x} and 1.2);
    \draw[nblue!30, thick, dashed] (0,-1.2) arc (270:90:{\x} and 1.2);
  }
  \draw[nblue!30, thick] (0,1.2) -- (0,-1.2);
  
  \draw[nblue!50, thick, opacity=0.8] (1.2,0) arc (0:-180:1.2 and 0.3);
  \draw[nblue!50, thick, dashed, opacity=0.8] (-1.2,0) arc (180:0:1.2 and 0.3);

  \draw[nblue!50, thick] (0,0) circle (1.2);

  \foreach \x/\y in {
    0.44/0.64, 0.54/0.70, 0.48/0.74, 0.58/0.62, 0.51/0.58,
    0.61/0.68, 0.42/0.71, 0.56/0.76, 0.46/0.61, 0.59/0.72,
    0.43/0.66, 0.52/0.59, 0.57/0.65, 0.49/0.78, 0.55/0.56
  } {
    \node[draw=ngreen!70!black, fill=ngreen!50, regular polygon, regular polygon sides=3, minimum size=3.5pt, inner sep=0pt, opacity=0.8] at (\x,\y) {};
  }

  \draw[ngreen!50, dashed, thick] (0.515,0.67) ellipse (0.13 and 0.12);

  \node[draw=nblue!70!black, fill=nblue, star, star points=5, minimum size=4pt,
        inner sep=0pt] at (0.50,0.68) {};
  \node[font=\tiny, nblue!80, anchor=south west] at (0.62,0.72) {anch.};

  \node[font=\scriptsize, ngreen!75] at (0,-1.5)
    {$\sigma_{\mathrm{ant}} \!\ll\! \sigma_{\mathrm{obs}}$};
\end{scope}

\draw[gray!30, thick] (6.8, 6.2) -- (6.8, 0.8);


\begin{axis}[
  at={(7.2cm, 0.8cm)},
  anchor=south west,
  width=9.5cm, height=6.2cm,
  title={\textbf{(c) UMAP Projection of Semantic Embeddings (UCF-Crime)}},
  title style={font=\bfseries\scriptsize, yshift=-1.5ex},
  xlabel={\scriptsize UMAP dim-1 (cosine metric)},
  ylabel={\scriptsize UMAP dim-2},
  xlabel style={font=\scriptsize},
  ylabel style={font=\scriptsize},
  xmin=-5.5, xmax=19.5,
  ymin=-0.5, ymax=13.5,
  grid=major, grid style={dashed, gray!15},
  tick style={thin, black},
  tick label style={font=\scriptsize},
  legend style={at={(0.02,0.98)}, anchor=north west, font=\tiny, row sep=-2pt, fill opacity=0.9, text opacity=1, draw=gray!30},
  legend cell align={left},
]

\addplot[
  only marks, mark=o,
  mark options={fill=nred!30, draw=nred, opacity=0.6},
  mark size=1.5pt,
] table[x=x, y=y, col sep=comma]
  {figures_ucf/data/umap_obs_ucf.csv};
\addlegendentry{Observation emb.\ ($\mathbf{s}^{\mathrm{obs}}$)}

\addplot[
  only marks, mark=triangle*,
  mark options={fill=ngreen!50, draw=ngreen!70!black, opacity=0.6},
  mark size=1.5pt,
] table[x=x, y=y, col sep=comma]
  {figures_ucf/data/umap_ant_ucf.csv};
\addlegendentry{Anticipation emb.\ ($\mathbf{s}^{\mathrm{sharp}}$)}

\addplot[
  only marks, mark=x,
  mark options={draw=norange, line width=1.2pt},
  mark size=2.5pt,
] table[x=x, y=y, col sep=comma]
  {figures_ucf/data/umap_cmp_ucf.csv};
\addlegendentry{Cross-Manifold Probe}

\addplot[
  only marks, mark=star,
  mark options={fill=nblue, draw=nblue!70!black},
  mark size=4.0pt,
] table[x=x, y=y, col sep=comma]
  {figures_ucf/data/umap_anchor_ucf.csv};
\addlegendentry{Text anchor}

\node[font=\scriptsize, nred!80, anchor=south]
  at (axis cs:15.35,8.7) {$\mathbf{s}^{\mathrm{obs}}$ cluster};
\draw[nred!45, dashed, thick] (axis cs:12.0,1.8) rectangle (axis cs:18.3,9.0);

\node[font=\scriptsize, ngreen!70!black, anchor=north]
  at (axis cs:-1.75,0.6) {$\mathbf{s}^{\mathrm{sharp}}$ manifold};
\draw[ngreen!45, dashed, thick] (axis cs:-5.0,0.8) rectangle (axis cs:1.6,7.3);

\node[font=\scriptsize, nblue!90, anchor=south west]
  at (axis cs:16.2,11.55) {text anchors};
\draw[nblue!40, dashed, thick] (axis cs:15.7,11.2) rectangle (axis cs:16.4,11.9);

\end{axis}
\end{tikzpicture}%
}
\vspace{-2ex}
\caption{\textbf{The Latent Clarity Hypothesis: conceptual geometry and empirical validation.}
(a)~Observation states ($\mathbf{s}^{\mathrm{obs}}$) are dispersed on the semantic manifold
($\mathcal{S}^{2047}$) because the underlying Observation Tensors ingest aleatoric noise
(motion blur, occlusion, camera shake) at the kinematic peak.
(b)~Anticipation embeddings ($\mathbf{s}^{\mathrm{sharp}}$) cluster tightly near the text
anchor because V-JEPA\,2's temporal predictor discards stochastic residuals, retaining
only the predictable physical trajectory.
(c)~UMAP projection (cosine metric)~\cite{umap} of a subset of real Semantic Latent States on UCF-Crime
($n{=}429$ obs, $n{=}429$ ant, $n{=}50$ CMP, $14$ anchors, drawn from the $1{,}193$-chunk validation set). While Anticipation embeddings
maintain consistently higher 2048-D cosine similarity to anchors (yielding superior zero-shot
accuracy), UMAP prioritizes local topology, revealing that Observation (red) and Anticipation
(green) embeddings occupy geometrically distinct sub-manifolds. Text anchors (blue stars)
form an isolated modality cluster. Cross-Manifold Probe points (orange) land completely in
the Anticipation region, confirming the ASP acts as a horizon-specific translation operator
($\Delta$) rather than a generic denoiser---an Asymmetric Geometric Overshoot that explains
the $7.3\%$ accuracy collapse (Section~\ref{sec:cmp}).}
\label{fig:teaser}
\label{fig:umap}
\end{figure*}

\paragraph{Notation.} \textbf{PULS} = KSD Bridge + ASP. The \textbf{Semantic
Manifold} is the unit sphere $\mathcal{S}^{2047} \subset \mathbb{R}^{2048}$
defined by \textsc{Qwen3-VL-Embedding-2B}~\cite{qwen3embed}; a \textbf{Semantic Latent State}
$\mathbf{s} \in \mathcal{S}^{2047}$ is the output of either the KSD Bridge
or the ASP.

\section{Related Work}
\label{sec:related}

\textbf{MIL-based VAD.}
Sultani et al.~\cite{sultani2018} established UCF-Crime and the MIL-ranking
recipe. Subsequent work~\cite{rtfm,s3r,mgfn,urdmu,dakd} pushed frame-level
AUROC from $75.4\%$ to $88.15\%$. All variants share three limitations:
(i) the output is a scalar with no semantics; (ii) detection is reactive
with no lead time; and (iii) the representation cannot support zero-shot VQA.

\textbf{Joint-Embedding Predictive Architectures.}
I-JEPA~\cite{ijepa} and V-JEPA~\cite{vjepa} learn predictable latent structure
rather than reconstructing pixels~\cite{lecun_path,mae}, discarding aleatoric noise
irrelevant to semantic reasoning. V-JEPA\,2~\cite{vjepa2} adds a temporal
predictor for action anticipation in closed vocabulary classification tasks. VL-JEPA~\cite{vljepa} extends JEPA to vision--language
by predicting continuous text embeddings; our KSD Bridge adapts this to stationary camera video anomaly sensing with
V-JEPA\,2 tokens. Concurrent V-JEPA\,2.1~\cite{vjepa21} alters the latent topology;
upgrading PULS requires re-distilling the KSD Bridge, left for future work.

\textbf{SALT distillation.}
Static-teacher Asymmetric Latent Training~\cite{salt} replaces the standard EMA
teacher with a frozen pretrained anchor and asymmetric noise-injection, avoiding
bi-level optimization~\cite{vicreg}. We apply SALT twice: aligning V-JEPA\,2 tokens
to the \textsc{Qwen3-VL} semantic manifold (KSD Bridge), and aligning
hallucinated-future embeddings to observed presents (ASP).

\textbf{Vision--language bridges.}
While large multimodal models~\cite{clip,qwen3embed,qwen3vl} achieve strong
anomaly classification, their autoregressive decoding induces latencies
incompatible with real-time surveillance. PULS sidesteps decoding: the KSD
Bridge produces a continuous semantic latent state in a single forward pass,
ensuring sub-frame latency on consumer hardware while retaining
zero-shot semantic tasks such as discriminative VQA, classification, retrieval and aid generative VQA with reasoning.

\section{KSD Bridge: Kinematic-to-Semantic Distillation}
\label{sec:ksd}

\paragraph{V-JEPA\,2 Token Geometry.} We use the frozen V-JEPA\,2 ViT-Large encoder~\cite{vjepa2}. A 16-frame, $256{\times}256$ RGB window (sampled at 30\,fps, stride 2) is tokenised into $8{\times}256 {=} 2048$ spatial-temporal tokens $\mathbf{Z} \in \mathbb{R}^{2048 \times 1024}$. The first 1024 tokens are the \emph{context}; the second 1024 are the \emph{target}. V-JEPA\,2's internal temporal predictor~\cite{vjepa2} autoregresses the target, yielding $\hat{\mathbf{Z}}_{\mathrm{future}} \in \mathbb{R}^{1024 \times 1024}$. Two tensors are available per clip: the Observation Tensor $\mathbf{Z}^{\mathrm{obs}} = [\mathbf{Z}_{\mathrm{ctx}} \,;\, \mathbf{Z}_{\mathrm{target}}]$ and the Anticipation Tensor $\mathbf{Z}^{\mathrm{ant}} = [\mathbf{Z}_{\mathrm{ctx}} \,;\, \hat{\mathbf{Z}}_{\mathrm{future}}]$.

\begin{figure*}[t]
\centering
\input{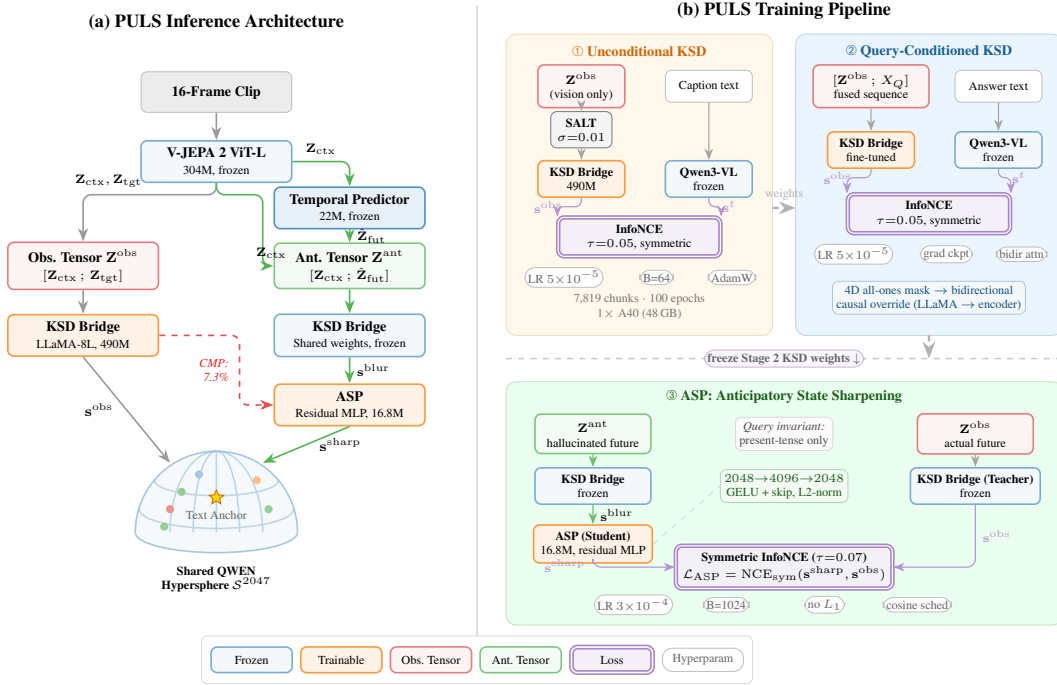}
\caption{\textbf{The Consolidated PULS Architecture and Training Strategy.} Raw video frames are encoded by the frozen V-JEPA 2 ViT-L. \textbf{(a) Left Track (Observation):} The context and target tokens form the Observation Tensor, yielding the observation embedding $\mathbf{s}^{\mathrm{obs}}$. During ASP training, this track acts as the frozen teacher. \textbf{(a) Right Track (Anticipation):} The Temporal Predictor hallucinates $\hat{\mathbf{Z}}_{\mathrm{fut}}$. The Anticipation Tensor yields a blurred state, which the trainable ASP (student) sharpens into $\mathbf{s}^{\mathrm{sharp}}$. \textbf{Objective:} A Symmetric InfoNCE loss (purple) explicitly distills the student's output toward the teacher's target, ensuring both tracks converge onto the \textbf{Shared Qwen Hypersphere} ($\mathcal{S}^{2047}$). The \textit{Cross-Manifold Probe} (dashed red line) explicitly misroutes the observation embedding into the ASP, triggering geometric collapse. \textbf{Right panel (b):} Three-stage training recipe --- \ding{192} unconditional caption alignment, \ding{193} query-conditioned VQA fine-tuning (both with SALT noise and frozen Qwen teacher), and \ding{194} ASP sharpening via frozen teacher distillation ($\tau{=}0.07$, batch 1024).}
\label{fig:arch}
\label{fig:puls}
\end{figure*}

\subsection{Architecture}
We instantiate $\pi_{\theta}$ with the last 8 layers of \textsc{LLaMA-3.2-1B}~\cite{llama32} (rotary embeddings~\cite{rope}, bidirectional attention). A per-token linear projection $W_{\mathrm{in}} \in \mathbb{R}^{1024 \times 2048}$ maps each V-JEPA\,2 token from $\mathbb{R}^{1024}$ to $\mathbb{R}^{2048}$. Given the projected token sequence $\mathbf{Z}W_{\mathrm{in}} \in \mathbb{R}^{2048 \times 2048}$ and query tokens $X_{Q} \in \mathbb{R}^{Q \times 2048}$ (embedded via \textsc{LLaMA}'s token embedding layer), concatenation along the sequence dimension yields a $(2048{+}Q) \times 2048$ input to $\pi_{\theta}$. The bridge outputs a Semantic Latent State via masked mean-pooling over non-PAD positions:
\begin{equation}
\mathbf{s} \;=\; \mathrm{L2Norm}\!\left(\mathrm{MeanPool}_{\mathrm{mask}}\big(\pi_{\theta}([\mathbf{Z}W_{\mathrm{in}} \,;\, X_{Q}])\big)\right) \;\in\; \mathcal{S}^{2047}.
\end{equation}
We write $f_{\theta}(\mathbf{Z}, X_{Q}) \equiv \mathrm{L2Norm}(\mathrm{MeanPool}_{\mathrm{mask}}(\pi_{\theta}([\mathbf{Z}W_{\mathrm{in}} \,;\, X_{Q}])))$ for the complete bridge and formally treat $\mathbf{s}$ as a dynamic \textbf{state vector} on the Semantic Manifold, enabling the ASP to perform temporal operations entirely in the latent domain.

\subsection{Training Objective}
We employ \emph{symmetric InfoNCE}~\cite{infonce,clip} ($\tau{=}0.05$). Following SALT~\cite{salt}, we inject Gaussian noise
$\epsilon \sim \mathcal{N}(0,\sigma^{2}\mathbf{I})$ with $\sigma{=}0.01$
into the V-JEPA\,2 tokens before the bridge during training; the Qwen text encoder is frozen
throughout. For a batch of $B$ pairs, let $\mathbf{s}_{i}^{\mathrm{obs}} = f_{\theta}(\mathbf{Z}_{i}^{\mathrm{obs}}{+}\epsilon_{i}, X_{Q})$ denote the KSD Bridge output on the noise-augmented Observation Tensor, and $\mathbf{s}_{i}^{t}$ the frozen \textsc{Qwen3-VL} embedding of the ground-truth QA caption. (Later, $\mathbf{s}^{\mathrm{blur}}$ and $\mathbf{s}^{\mathrm{sharp}}$ denote ASP operations on the Anticipation Tensor $\mathbf{Z}^{\mathrm{ant}}$).
\begin{equation}
\mathcal{L}_{\mathrm{KSD}} = -\frac{1}{2B}\sum_{i=1}^{B}
  \left[
    \log\frac{e^{\langle \mathbf{s}_{i}^{\mathrm{obs}},\mathbf{s}_{i}^{t} \rangle /\tau}}
             {\sum_{j} e^{\langle \mathbf{s}_{i}^{\mathrm{obs}},\mathbf{s}_{j}^{t} \rangle /\tau}}
   +\log\frac{e^{\langle \mathbf{s}_{i}^{t},\mathbf{s}_{i}^{\mathrm{obs}} \rangle /\tau}}
             {\sum_{j} e^{\langle \mathbf{s}_{i}^{t},\mathbf{s}_{j}^{\mathrm{obs}} \rangle /\tau}}
  \right].
\end{equation}
Batches of size 64 are processed on a \emph{single}
NVIDIA A40 (48\,GB) GPU.

To prevent catastrophic causal masking, we explicitly override the LLaMA default with a bidirectional attention mask (see Appendix~\ref{app:bidir}).

\subsection{Training Data Curation}
\textbf{Kinematic-peak filtering.} UCF-Crime's temporal annotations~\cite{sultani2018,ucf_annot} define
macro-windows (often $>30$\,s) while the actual kinematic event spans
$\sim$1--2\,s. We run a sliding-window optical-flow-magnitude scan and retain
the single window of highest variance per annotated segment as the \emph{Peak}
chunk. From 13 UCF-Crime anomaly categories plus Normal, this yields
$7{,}819$ curated 1-second training chunks.

\textbf{Semantic stripping.} QA captions (10 per chunk, generated by \textsc{Qwen2.5-7B}) describe kinematics and severity, never lighting, clothing, or static scene cues---explicitly mitigating static representation bias (the colloquial ``white-shirt'' shortcut)~\cite{li2018resound}.

\subsection{KSD Bridge Validation}
\label{sec:ksd-results}
On 1{,}150 held-out UCF-Crime chunks, the KSD Bridge achieves chunk-level AUROC $\mathbf{0.8994}$ (via cosine distance against danger-weighted prototypes) and $34.9\%$ mean zero-shot VQA accuracy. Full-set evaluation on 290 untrimmed videos via Savitzky--Golay temporal broadcasting~\cite{savgol} yields frame-level AUROC $0.6282$, reflecting a structural penalty incurred by predictive (vs.\ reactive) detectors (see Section~\ref{sec:protocols}, Appendix~\ref{app:hparams} and Table~\ref{tab:cross_dataset}).

\subsection{Cross-Manifold Probe: The ASP is a Horizon-Specific Operator}
\label{sec:cmp}
As a decisive negative control, the Cross-Manifold Probe evaluates the ASP on the \emph{Observation} Tensor: $\mathbf{s}^{\mathrm{cmp}} = g_\phi(f_\theta(\mathbf{Z}^{\mathrm{obs}}, X_Q))$. On the 1{,}193-chunk validation set, Q1 accuracy collapses to $7.29\%$ (below random chance), binary AUROC drops to $0.5535$, and per-task scores plummet. We term this \textbf{Asymmetric Geometric Overshoot}: the ASP applies a delta calibrated for \emph{blurred} future latents; on an already-sharp observation latent, it overshoots into unoccupied hypersphere regions (confirmed by UMAP~\cite{umap}, Fig.~\ref{fig:teaser}c). This collapse proves the ASP is not merely a generic classifier head, but a horizon-conditional operator (see Section~\ref{sec:topological-sinks} for geometric analysis and Section~\ref{sec:cmp-analysis} for a per-category breakdown).

\section{PULS Pipeline: Anticipatory State Predictor}
\label{sec:asp}

\subsection{Motivation: Epistemic blur}
The temporal predictor's hallucination is deterministic (MSE-trained) and
therefore outputs the conditional mean over plausible futures --- a
``blurred'' latent. Passing $\mathbf{Z}^{\mathrm{ant}}$ through the frozen KSD
Bridge yields a semantic vector $\mathbf{s}^{\mathrm{blur}}$ that is
approximately centred on, but not identical to, the observation vector
$\mathbf{s}^{\mathrm{obs}}$ derived from the actual future.

\subsection{The Anticipatory State Predictor}
We train a lightweight 2-layer residual MLP $g_{\phi}: \mathbb{R}^{2048} \to \mathbb{R}^{2048}$. Maintaining the row-vector convention of the Semantic Manifold, the projection is defined as:
\begin{equation}
g_{\phi}(\mathbf{s}) = \mathrm{L2Norm}\!\left(\mathbf{s} + \mathrm{GELU}(\mathbf{s}W_{1} + \mathbf{b}_{1})W_{2} + \mathbf{b}_{2}\right),
\end{equation}
where $W_{1}\in\mathbb{R}^{2048\times 4096}$ and $W_{2}\in\mathbb{R}^{4096\times 2048}$ ($\approx$16.8M parameters). During training, the KSD Bridge and both tensor extraction paths remain strictly frozen. The geometric expansion--contraction $2048{\to}4096{\to}2048$ disentangles mixed future-probability modes before contractive projection back to the unit hypersphere.

\subsection{Loss choice and the collapse of $L_{1}$}
While SALT~\cite{salt} prescribes $L_1$ latent regression, in our setting $L_1$ collapsed to a low-magnitude mean vector, dropping 14-way accuracy to $34.2\%$. We replace $L_1$ with \emph{symmetric InfoNCE}~\cite{infonce}. For a batch of $B=1024$ pairs, the loss is defined over the empirical batch expectation:
\begin{equation}
\mathcal{L}_{\mathrm{ASP}} = \frac{1}{B} \sum_{i=1}^{B} \mathcal{L}_{\mathrm{NCE}}^{\mathrm{sym}}\!\left(g_{\phi}(\mathbf{s}_{i}^{\mathrm{blur}}),\; \mathbf{s}_{i}^{\mathrm{obs}} ;\; \tau{=}0.07\right).
\end{equation}
InfoNCE's uniformity penalty forces the output to span the semantic manifold, restoring angular class structure despite a minor AUROC regression ($0.8994 \to 0.8680$)---a deliberate trade given discriminativity is an angular property on the unit sphere.

\subsection{Query conditioning and temporal invariance}
While the KSD Bridge is query-conditioned for open-vocabulary VQA, ASP training requires strict geometric isolation. Mismatched or future-tense queries (e.g., ``What will happen next?'') trigger a temporal \emph{double-shift}: V-JEPA\,2 has already advanced tokens into the physical future, causing the bridge to forecast $T{+}2$ semantics from $T{+}1$ physics. To prevent manifold misalignment, a strict present-tense query invariant ($X_{Q}=$ ``What is the primary action and severity?'') is injected into both paths during ASP distillation. This guarantees the ASP strictly learns to correct the epistemic blur of hallucinated physical geometry, while inheriting the KSD Bridge's VQA generalisation at inference.

\section{Evaluation Protocols and Datasets}
\label{sec:protocols}

\subsection{UCF-Crime}
\label{sec:ucf-dataset}
We evaluate our framework using the UCF-Crime dataset, a large-scale surveillance benchmark containing 13 anomaly categories plus Normal~\cite{sultani2018}. Rather than training on the raw full-length video distribution, our models are trained on a curated kinematic-peak filtered subset of the UCF-Crime training split, and evaluated on held-out validation splits as well as on out-of-distribution (OOD) transfer to the XD-Violence dataset. We structure the dataset splits as follows:

\textbf{Training Split.} The KSD Bridge and ASP are trained exclusively on the UCF-Crime training split, consisting of 1{,}401 videos. Applying the kinematic-peak filtering protocol (detailed in Section~\ref{sec:ksd}) yields $\mathbf{7{,}819}$ high-fidelity kinematic-peak chunks (1\,s sliding window with 0.5\,s stride). Each chunk produces two core tensor artifacts: an observation tensor $[\mathbf{Z}_{\mathrm{ctx}} \,;\, \mathbf{Z}_{\mathrm{target}}]$ and an anticipation tensor $[\mathbf{Z}_{\mathrm{ctx}} \,;\, \hat{\mathbf{Z}}_{\mathrm{fut}}]$, both represented in bfloat16 precision $\in \mathbb{R}^{2048\times 1024}$ of V-JEPA\,2 ViT-L spatial-temporal tokens.

\textbf{Evaluation Splits.} Rather than relying on a single aggregate metric, we construct three non-overlapping evaluation splits from the held-out distribution to isolate different facets of the model's performance:
\begin{itemize}[leftmargin=*,nosep]
  \item \textbf{A1 --- Frame-level AUROC} (290 videos, 140 anomaly / 150 normal): The official test split from Sultani et al.~\cite{sultani2018} containing frame-level temporal annotations. This enables direct comparison against traditional Weakly-Supervised MIL baselines (e.g., RTFM, MGFN, UR-DMU).
  \item \textbf{A2 --- Semantic VQA} (1{,}193 chunks, 413 videos): A chunk-level validation set containing 596 anomaly and 597 normal chunks balanced across all 14 categories. The 1{,}150 chunks reported in Table~\ref{tab:vqa} represent the strictly held-out test+val intersection, while the full 1{,}193 set contains the expanded validation distribution used to verify the Cross-Manifold Probe's geometric collapse.
  \item \textbf{A3 --- Lead-Time Anticipation} (561 of 596 untrimmed anomaly videos): Used to evaluate the model's temporal anticipation horizons. We exclude 35 anomaly videos from the validation set due to missing start-frame annotations which are required to align temporal offsets relative to the anomaly onset $T_0$.
\end{itemize}
The denominators differ between these splits due to metadata constraints: Frame-Level AUROC (A1) requires frame-level ground-truth annotations (available only for the 290 test videos), whereas Semantic VQA (A2) relies on video-level labels for 14-way chunk classification. Lead-time anticipation (A3) is restricted to the 561 untrimmed anomaly videos with annotated physical onset times, since normal clips have no onset.

\subsection{XD-Violence (Out-of-Distribution)}
\label{sec:xd-dataset}
XD-Violence contains 7 anomaly categories (Explosion, Shooting, Riot, Fighting, CarAccident, Abuse, Normal) under both static and dynamic (ego-motion) cameras~\cite{xd_violence}. To verify out-of-distribution (OOD) zero-shot transfer, we evaluate on a held-out set of 1{,}200 chunks (600 anomaly / 600 normal) and frame-level metrics on 800 videos \emph{without any retraining} (weights are frozen as trained on UCF-Crime). The audio channel is omitted, and license restrictions are respected (research use only).

\subsection{Discriminative VQA Protocol}
\label{sec:vqa-protocol}
We build a prototype cache of $5\times 14 = 70$ answer embeddings (5 VQA tasks $\times$ 14 categories) via \textsc{Qwen3-VL-Embedding-2B}. A chunk is classified per task by $\operatorname*{arg\,max}_{c} \langle \mathbf{s}, \mathbf{a}_{q,c}\rangle$; we report per-task accuracy, mean across 5 tasks, and majority-vote category. To compute binary AUROC, we project the similarities into a scalar anomaly score, defined formally as $y(\mathbf{s}) = \max_{q,c} \big( \langle \mathbf{s}, \mathbf{a}_{q,c} \rangle \cdot w_c \big)$, where $w_c \in [0,1]$ is a deterministic danger weight encoding kinetic severity (e.g., Normal=0.0, Shoplifting=0.5, Explosion=1.0). Crucially, these weights are formulated as \textbf{ontological constants}—representing the inherent physical severity of an event independent of any specific dataset. They are applied identically across both evaluations (UCF-Crime, XD-Violence) without target-domain calibration, ensuring absolute integrity in our zero-shot and out-of-distribution testing protocols. Legacy MIL methods perform an equivalent collapse implicitly via their binary output head; our formulation makes this projection explicit and interpretable.

\section{Results}
\label{sec:results}

\subsection{VQA Results: KSD Bridge vs.\ ASP on UCF-Crime}
\label{sec:vqa}

\begin{table}[t]
\caption{\textbf{14-way zero-shot VQA on UCF-Crime.} KSD Bridge and ASP evaluated on 1{,}150 held-out chunks. The Cross-Manifold Probe (Observation Tensor $\to$ KSD Bridge $\to$ ASP) is evaluated on the 1{,}193-chunk validation set. ASP dominates KSD Bridge across all tasks despite consuming \emph{hallucinated} futures. The Cross-Manifold Probe's collapse to random chance demonstrates ASP is a horizon-specific operator (Section~\ref{sec:cmp}).}
\label{tab:vqa}
\centering
\small
\begin{tabular}{l c c c c}
\toprule
\textbf{Task} & \textbf{KSD Bridge (obs.)} & \textbf{Cross-Manifold Probe} & \textbf{ASP (ant.)} & $\Delta$(ASP$-$KSD) \\
\midrule
Q1 --- Predictive anticipation       & 28.7\% & \phantom{0}7.3\% & \textbf{42.7\%} & $+14.0$ \\
Q2 --- Incident type                 & 55.6\% & 16.7\% & \textbf{55.7\%} & $+\phantom{0}0.1$ \\
Q3 --- Hazard assessment             & 23.2\% & \phantom{0}4.8\% & \textbf{34.4\%} & $+11.2$ \\
Q4 --- Activity recognition          & 35.3\% & \phantom{0}6.0\% & \textbf{46.4\%} & $+11.1$ \\
Q5 --- Severity estimation           & 31.6\% & \phantom{0}5.2\% & \textbf{43.3\%} & $+11.6$ \\
\midrule
\textbf{Mean across tasks}           & 34.9\% & \phantom{0}8.0\% & \textbf{44.5\%} & $+\phantom{0}9.6$ \\
Majority-vote 14-way                 & 37.1\% & \phantom{0}7.3\% & 34.2\%$^{\dagger}$ & $-\phantom{0}2.9$ \\
Binary AUROC                         & 0.8994 & 0.5535 & 0.8680 & $-0.0314$ \\
\bottomrule
\end{tabular}\\[2pt]
{\footnotesize $^{\dagger}$ Under the symmetric InfoNCE objective, the Semantic Manifold is strictly $L_2$-normalized. While this mathematically optimizes the per-task angular separation (driving the $+9.6$\,pp mean gain), it intentionally collapses the scalar magnitude variances that naive majority-voting relies upon, resulting in a minor regression.}
\end{table}

Table~\ref{tab:vqa} reports per-task results. \textbf{(i)~Kinematic gains:} Predictive (+14.0pp), severity (+11.6pp), hazard (+11.2pp), and activity (+11.1pp) improve substantially. Q2 (incident type) is unchanged ($+0.1$pp), as coarse labels are decided at the context level and do not require future-physics reasoning. \textbf{Baselines:} Under the same protocol, VL-JEPA~\cite{vljepa} scores $19.5\%$ (Gemma-300M) to $29.6\%$ (Qwen3-8B), and C3D~\cite{c3d,sultani2018} scores $23.0\%$. ASP's $44.5\%$ mean exceeds the strongest VL-JEPA ablation by $+14.9$pp, without additional supervision. \textbf{(ii)~AUROC--angular trade:} Binary AUROC drops ($0.8994 \to 0.8680$) because InfoNCE $L_{2}$-normalises outputs, collapsing scalar magnitude. Against a $14$-way fine-grained manifold, angular separation is the correct property to optimise. 

\textbf{Frame-level AUROC.} Full-set evaluation over 290 untrimmed UCF-Crime
test videos yields frame-level AUROC $\mathbf{0.6282}$. Per-category variance
validates Semantic Stripping: kinematic-dense categories localize well
(Explosion $0.795$, RoadAccidents $0.788$, Fighting $0.743$), while
appearance-based categories fall to chance (Shoplifting $0.400$, Burglary
$0.412$). MIL methods exploit static scene regularities in these low-kinematics
categories; PULS ignores non-physical cues by design, incurring the anticipated
metric penalty. Frame-level AUC additionally penalizes any architecture that
anticipates anomalies before the reactive ground-truth mask begins.

\subsection{Out-of-Distribution Evaluation: XD-Violence}
\label{sec:xd}

\begin{table}[ht]
\caption{\textbf{Cross-Dataset Evaluation.} Comparison of in-distribution (UCF) vs.\ OOD (XD) metrics, evaluating the KSD Bridge (v8) and ASP (v9). Anticipation Advantage measured at $T_{\mathrm{onset}}$.}
\label{tab:cross_dataset}
\centering
\small
\begin{tabular}{lccc}
\toprule
\textbf{Metric} & \textbf{UCF-Crime (In-Dist)} & \textbf{XD-Violence (OOD)} & \textbf{Retention} \\
\midrule
Frame-level AUROC & 0.6282 & 0.7894 & --- \\
Chunk AUROC (KSD Bridge) & 0.8994 & 0.8162 & 91\% \\
Chunk AUROC (ASP) & 0.8680 & 0.7003 & 81\% \\
Mean 5Q VQA (Observation) & 34.9\% & 27.9\% & 80\% \\
Mean 5Q VQA (Anticipation) & 44.5\% & 27.4\% & 62\% \\
$\Delta$ (Anticipation $-$ Obs.) & $+9.6$pp & $-0.5$pp & --- \\
Anticipation Advantage ($T_{\mathrm{onset}}$) & $+3.7$pp & $+3.0$pp & --- \\
\bottomrule
\end{tabular}
\end{table}

\textbf{Metrics \& Distillation Bias.} Table~\ref{tab:cross_dataset} details the global OOD transfer metrics. The ASP chunk-AUROC regression relative to UCF-Crime ($-16.8$pp) traces to a dataset-induced distillation bias: the ASP was distilled against a static-camera teacher. XD-Violence's dynamic cameras produce ego-motion residuals that the surprise gate correctly elevates, but the ASP misinterprets as semantic kinematic escalation. Retraining on a mixed-camera corpus is the direct fix.

\textbf{Newtonian-invariant categories.} Despite the distribution shift, categories whose kinematic signatures are camera-motion-agnostic improve: CarAccident $+24.0$pp ($p{<}0.001$, McNemar) and Explosion $+1.7$pp Q1 accuracy relative to the KSD Bridge. Crucially, open-vocabulary generalization is preserved: the KSD Bridge achieves $\mathbf{69.5\%}$ zero-shot Q1 accuracy on ``Riot''---a category entirely absent from the UCF-Crime training distribution---before degrading to $48.3\%$ under the ASP due to the aforementioned ego-motion distillation bias.

\textbf{Triple-Track on XD-Violence.} The peak gated--context advantage is $+4.5$pp at $T{-}1.5$s and $+3.0$pp at $T_0$ ($p{<}0.001$ at all horizons $T_0$--$T{-}2.0$s, $N{=}1000$ permutation), consistent with the longer kinematic escalation windows typical of vehicular and crowd-dynamics events.

\textbf{Latent Clarity Hypothesis: Claim 1 vs.\ Claim 2.}
The avg-5Q VQA accuracy is flat between KSD Bridge ($27.9\%$) and ASP ($27.4\%$)
on XD-Violence, isolating two separable claims. \emph{Claim 1}: the anticipation
tensor encodes domain-agnostic predictive structure. \emph{Claim 2}: the ASP
sharpener decodes it effectively. The OOD lead-time advantage at $T_0$
($+3.0$pp, $p{<}0.001$) confirms Claim 1 is OOD-robust --- the hallucinated
future natively encodes incident-relevant kinematics regardless of camera domain.
The flat avg-5Q indicates Claim 2 is domain-specific: the ASP was distilled on
UCF-Crime's static-camera manifold and has not seen XD's ego-motion geometry.
Per-class evidence is mechanistically coherent: Newtonian-invariant categories
improve (Explosion avg-5Q $+10.2$pp, CarAccident $+9.2$pp) while
ego-motion-dominated categories degrade (Fighting $-22.2$pp, Shooting $-8.1$pp)
--- exactly where UCF-to-XD domain shift is maximal.

\subsection{Cross-Manifold Probe: Per-Category Analysis}
\label{sec:cmp-analysis}

To verify the topological alignment (and crucial separation) between the Observation and Anticipation sub-manifolds, we conduct a fine-grained category-by-category analysis. Table~\ref{tab:per_category} details the zero-shot Q1 VQA accuracy across all categories for the Observation ($\mathbf{s}^{\mathrm{obs}}$), Anticipation ($\mathbf{s}^{\mathrm{sharp}}$), and Cross-Manifold Probe (CMP: $\mathbf{s}^{\mathrm{obs}}\to\mathrm{ASP}$) configurations on both the UCF-Crime validation set (1{,}193 chunks) and the XD-Violence transfer set (1{,}200 chunks).

\begin{table}[h]
\caption{\textbf{Per-Category VQA Accuracy (Q1) across Manifolds.} Actual results from the validation caches for Observation ($\mathbf{s}^{\mathrm{obs}}$), Anticipation ($\mathbf{s}^{\mathrm{sharp}}$), and the Cross-Manifold Probe ($\mathbf{s}^{\mathrm{obs}}\to\mathrm{ASP}$). All values are Q1 zero-shot accuracy. The collapse of these categories in the CMP columns confirms the topological separation of the Anticipation and Observation sub-manifolds within the Shared Semantic Hypersphere. UCF results are In-Distribution; XD results are OOD Transfer.}
\label{tab:per_category}
\centering
\small
\begin{tabular}{l ccc ccc}
\toprule
& \multicolumn{3}{c}{\textbf{UCF-Crime (In-Distribution)}} & \multicolumn{3}{c}{\textbf{XD-Violence (OOD Transfer)}} \\
\cmidrule(r){2-4} \cmidrule(l){5-7}
\textbf{Category} & \textbf{Obs.} & \textbf{Ant.} & \textbf{CMP} & \textbf{Obs.} & \textbf{Ant.} & \textbf{CMP} \\
\midrule
Abuse         & 32.5\% & 7.5\%  & 2.5\%  & 16.7\% & 8.3\%  & 100\%$^{\dagger}$ \\
Arrest        & 82.1\% & 79.5\% & 0.0\%  & ---    & ---    & ---    \\
Arson         & 11.8\% & 50.0\% & 0.0\%  & ---    & ---    & ---    \\
Assault       & 0.0\%  & 8.3\%  & 0.0\%  & ---    & ---    & ---    \\
Burglary      & 34.4\% & 35.9\% & 6.3\%  & ---    & ---    & ---    \\
CarAccident   & 23.3\% & 55.8\% & 7.0\%  & 16.2\% & 40.2\% & 0.9\%  \\
Explosion     & 62.5\% & 67.5\% & 0.0\%  & 51.7\% & 53.4\% & 0.0\%  \\
Fighting      & 67.9\% & 73.6\% & 0.0\%  & 41.9\% & 40.2\% & 0.9\%  \\
Riot          & ---    & ---    & ---    & 69.5\% & 48.3\% & 0.0\%  \\
Robbery       & 36.5\% & 54.1\% & 66.2\% & ---    & ---    & ---    \\
Shooting      & 40.0\% & 24.0\% & 0.0\%  & 0.8\%  & 0.0\%  & 0.0\%  \\
Shoplifting   & 97.9\% & 97.9\% & 19.1\% & ---    & ---    & ---    \\
Stealing      & 13.5\% & 5.4\%  & 0.0\%  & ---    & ---    & ---    \\
Vandalism     & 53.8\% & 61.5\% & 0.0\%  & ---    & ---    & ---    \\
Normal        & 14.4\% & 37.4\% & 0.0\%  & 4.2\%  & 6.2\%  & 2.2\%  \\
\midrule
\textbf{Mean (Q1)} & \textbf{28.7\%} & \textbf{42.7\%} & \textbf{5.5\%} & \textbf{19.9\%} & \textbf{21.0\%} & \textbf{2.3\%} \\
\bottomrule
\end{tabular}
\par\vspace{2pt}\footnotesize $^{\dagger}$Abuse in XD-Violence contains only 12 samples; results are statistically sensitive to scene priors.
\end{table}

\section{Triple-Track Lead-Time Protocol}
\label{sec:leadtime}

For each of 561 untrimmed UCF-Crime test videos with annotated onset
$T_{\mathrm{onset}}$, we extract a 16-frame window ending at
$T_{\mathrm{onset}}-\Delta$ for
$\Delta \in \{0, 0.5, 1.0, 1.5, 2.0, 3.0, 5.0\}$ seconds and run each window
through the PULS pipeline. Three tracks are evaluated:
\begin{description}[leftmargin=*,nosep]
  \item[\textbf{Track 1 --- Ungated anticipation.}] Evaluates the Anticipation
  Tensor $[\mathbf{Z}_{\mathrm{ctx}} \,;\, \hat{\mathbf{Z}}_{\mathrm{future}}]$ via the
  ASP without $L_1$-surprise gating, reporting raw 14-way majority-vote accuracy.
  \item[\textbf{Track 2 --- $L_{1}$-surprise gated anticipation.}] Let
  $L_{1}(t)=\|\hat{\mathbf{Z}}_{\mathrm{future}}(t)-\mathbf{Z}_{\mathrm{target}}(t)\|_{1}$
  be the V-JEPA\,2 prediction error at $t$. Compute a calm-zone baseline
  $(\mu_{c},\sigma_{c})$ from horizons $\Delta\!\geq\!3.0$s. Open a gate iff
  $L_{1}(T_{\mathrm{onset}}-\Delta) > \mu_{c}+z\sigma_{c}$ with $z=2.0$.
  Track 2 reports accuracy on gate-open windows only.
  \item[\textbf{Track 3 --- Context-only control.}] Replace the Anticipation
  Tensor with $[\mathbf{Z}_{\mathrm{ctx}} \,;\, \mathbf{Z}_{\mathrm{ctx}}]$ and
  apply the same ASP. This isolates scene bias.
\end{description}

\begin{table}[t]
\caption{\textbf{Triple-Track Lead-Time on UCF-Crime (561 videos).}
Track 2 decays monotonically as the anticipatory horizon extends, saturating cleanly
at $0\%$ in the calm zone, while Track 1 is horizon-invariant --- the
signature of scene-bias classification.}
\label{tab:leadtime}
\centering
\small
\begin{tabular}{l c c c c c}
\toprule
\textbf{Horizon} & \textbf{T1 Ungated} & \textbf{T2 Gated $z{=}2$} & \textbf{T3 Ctx-only} & \textbf{Gates opened} & \textbf{$\Delta$(T2$-$T3)} \\
\midrule
$T_{0}$    & 31.9\% & \textbf{37.0\%} & 33.3\% & 108 / 561 & $+3.7$ \\
$T{-}0.5$  & 32.9\% & \textbf{36.5\%} & 27.6\% & 123 / 554 & $\mathbf{+8.9}$ \\
$T{-}1.0$  & 33.3\% & \textbf{35.9\%} & 30.2\% & 106 / 544 & $+5.7$ \\
$T{-}1.5$  & 34.4\% & 34.6\%          & 31.9\% & 113 / 541 & $+2.7$ \\
$T{-}2.0$  & 32.5\% & 29.3\%          & 26.7\% & 116 / 538 & $+2.6$ \\
$T{-}3.0$  & 32.8\% & $0.0\%$ (calm)  & ---    & 0 / 528   & n/a \\
$T{-}5.0$  & 33.5\% & $0.0\%$ (calm)  & ---    & 0 / 508   & n/a \\
\bottomrule
\end{tabular}
\end{table}

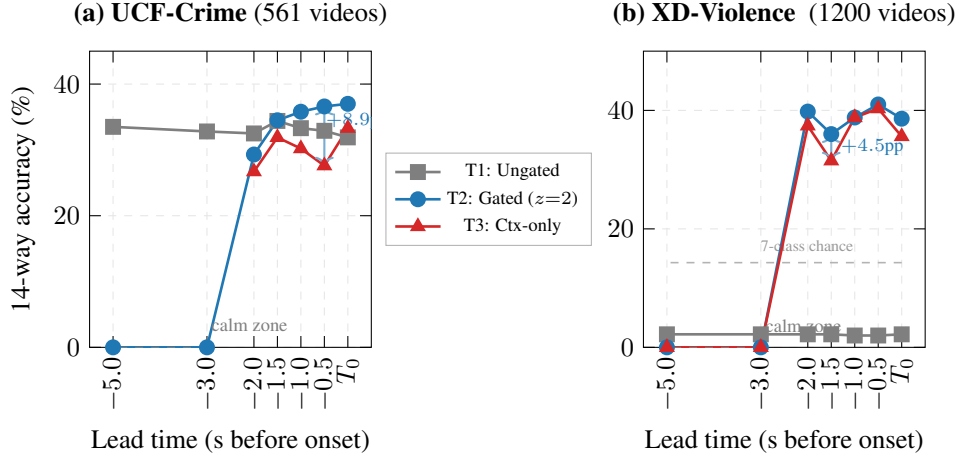
\begin{figure}[t]
\centering

\definecolor{nblue}{RGB}{31,119,180}
\definecolor{nred}{RGB}{214,39,40}
\definecolor{ngreen}{RGB}{44,160,44}
\definecolor{ngray}{RGB}{127,127,127}

\begin{tikzpicture}

\begin{axis}[
  name=ucf,
  width=0.38\linewidth, height=5.5cm,
  title={\textbf{(a) UCF-Crime} (561 videos)},
  xlabel={Lead time (s before onset)},
  ylabel={14-way accuracy (\%)},
  xmin=-0.5, xmax=5.5,
  ymin=0, ymax=45,
  xtick={0,0.5,1.0,1.5,2.0,3.0,5.0},
  xticklabels={$T_0$,$-0.5$,$-1.0$,$-1.5$,$-2.0$,$-3.0$,$-5.0$},
  x tick label style={rotate=90, anchor=east},
  x dir=reverse,
  grid=major, grid style={dashed, gray!20},
  legend style={font=\scriptsize, at={(1.05,0.5)}, anchor=west,
                draw=black!30, fill=white, legend columns=1},
  tick style={thin, black},
  every axis plot/.append style={line width=1.0pt, mark size=2.5pt},
]
\addplot[color=ngray, mark=square*, mark options={fill=ngray}] coordinates {
  (0,31.9)(0.5,32.9)(1.0,33.3)(1.5,34.4)(2.0,32.5)(3.0,32.8)(5.0,33.5)
};
\addlegendentry{T1: Ungated}

\addplot[color=nblue, mark=*, mark options={fill=nblue}] coordinates {
  (0,37.0)(0.5,36.6)(1.0,35.8)(1.5,34.5)(2.0,29.3)(3.0,0.0)(5.0,0.0)
};
\addlegendentry{T2: Gated ($z{=}2$)}

\addplot[color=nred, mark=triangle*, mark options={fill=nred}] coordinates {
  (0,33.3)(0.5,27.6)(1.0,30.2)(1.5,31.9)(2.0,26.7)
};
\addlegendentry{T3: Ctx-only}

\draw[<->, nblue!60, thick] (axis cs:0.5,27.6) -- (axis cs:0.5,36.6)
  node[midway, right, font=\scriptsize, nblue, xshift=-3pt, yshift = 5.7pt] {$+8.9$pp};

\draw[dashed, ngray] (axis cs:3.0,0) -- (axis cs:5.0,0);
\node[font=\scriptsize, ngray, anchor=south west] at (axis cs:3.1,1.0) {calm zone};

\end{axis}

\begin{axis}[
  name=xd,
  at={(ucf.east)}, anchor=west, xshift=3.6cm,
  width=0.38\linewidth, height=5.5cm,
  title={\textbf{(b) XD-Violence } (1200 videos)},
  xlabel={Lead time (s before onset)},
  xmin=-0.5, xmax=5.5,
  ymin=0, ymax=50,
  xtick={0,0.5,1.0,1.5,2.0,3.0,5.0},
  xticklabels={$T_0$,$-0.5$,$-1.0$,$-1.5$,$-2.0$,$-3.0$,$-5.0$},
  x tick label style={rotate=90, anchor=east},
  x dir=reverse,
  grid=major, grid style={dashed, gray!20},
  tick style={thin, black},
  every axis plot/.append style={line width=1.0pt, mark size=2.5pt},
]
\addplot[color=black!30, dashed, line width=0.6pt, forget plot] coordinates {
  (0,14.3)(5.0,14.3)
};
\node[font=\tiny, text=black!40, anchor=south west] at (axis cs:3.2,14.7) {7-class chance};

\addplot[color=ngray, mark=square*, mark options={fill=ngray}] coordinates {
  (0,2.2)(0.5,2.0)(1.0,2.0)(1.5,2.2)(2.0,2.2)(3.0,2.2)(5.0,2.2)
};

\addplot[color=nblue, mark=*, mark options={fill=nblue}] coordinates {
  (0,38.6)(0.5,41.0)(1.0,38.8)(1.5,36.0)(2.0,39.8)(3.0,0.0)(5.0,0.0)
};

\addplot[color=nred, mark=triangle*, mark options={fill=nred}] coordinates {
  (0,35.6)(0.5,40.3)(1.0,38.8)(1.5,31.5)(2.0,37.4)(3.0,0.0)(5.0,0.0)
};

\draw[<->, nblue!60, thick] (axis cs:1.5,31.5) -- (axis cs:1.5,36.0)
  node[midway, right, font=\scriptsize, nblue] {$+4.5$pp};

\draw[dashed, ngray] (axis cs:3.0,0) -- (axis cs:5.0,0);
\node[font=\scriptsize, ngray, anchor=south west] at (axis cs:3.1,1.0) {calm zone};

\end{axis}

\end{tikzpicture}
\caption{\textbf{Triple-Track Lead-Time Protocol.}
(a)~UCF-Crime: Track 2 (gated) decays monotonically as the horizon extends and saturates at $0\%$
in the calm zone ($T{-}3$s), while Track 1 (ungated) is flat --- the signature of
scene-bias classification. Peak $+8.9$pp gated--context advantage at $T{-}0.5$s.
(b)~XD-Violence (no retraining): Track 1 collapses to $2.2\%$ (below 7-class
chance of $14.3\%$) due to distribution shift, while gated Track 2 recovers
to $38$--$41\%$, demonstrating that the $L_1$-surprise gate compensates for
OOD scene bias. Peak gated--context advantage $+4.5$pp at $T{-}1.5$s.}
\label{fig:leadtime}
\end{figure}

\textbf{Double dissociation.}
Table~\ref{tab:leadtime} reveals the signature double dissociation: Track 1 is \emph{flat} ($31.9\%$--$34.4\%$ across all horizons including the impossible $T{-}5$s), confirming that raw ungated accuracy reflects static scene priors. Track 2 decays \emph{monotonically as the anticipatory horizon extends} ($37.0\% \to 29.3\%$), saturating cleanly to $0\%$ in the calm zone. The peak Track 2--Track 3 advantage of $\mathbf{+8.9}$pp at $T{-}0.5$s is the paper's central empirical claim.

\textbf{Permutation test ($N{=}1000$).}
To rule out confounding from class-imbalance selection by the surprise gate, we shuffle category labels across 561 videos 1000 times with surprise traces held fixed. Observed accuracies lie $>10\sigma$ above the null at all active horizons ($p{<}0.001$ throughout). The label-permutation null distribution results are detailed in Table~\ref{tab:perm}. Observed Track~2 accuracy is $\sim 4\times$ the null mean at all active horizons, proving that the gated anticipatory performance is statistically significant and cannot be explained by random selection or class-imbalance biases (where the random chance baseline is $\approx 1/14 \approx 7.1\%$).

\begin{table}[!ht]
\caption{\textbf{Label-permutation null distribution} ($N{=}1000$, chance $\approx 1/14 \approx 7.1\%$). Observed Track~2 accuracy is $\sim 4\times$ the null mean at all active horizons.}
\label{tab:perm}
\centering\small
\begin{tabular}{l c c c c}
\toprule
\textbf{Horizon} & \textbf{Observed} & \textbf{Null mean $\pm$ std} & \textbf{$p$} & \textbf{Sig.} \\
\midrule
$T_{0}$    & 37.0\% & $8.3\%\pm 2.6\%$ & $<0.001$ & \ding{51} \\
$T{-}1.0$  & 35.8\% & $8.2\%\pm 2.6\%$ & $<0.001$ & \ding{51} \\
$T{-}2.0$  & 29.3\% & $8.0\%\pm 2.5\%$ & $<0.001$ & \ding{51} \\
$T{-}3.0$  & 0.0\%  & 0.0\%            & 1.0 (calm) & --- \\
\bottomrule
\end{tabular}
\end{table}

\section{Discussion}
\label{sec:discuss}

\subsection{The Latent Clarity Hypothesis}
Table~\ref{tab:vqa} quantifies a counter-intuitive inequality: for kinematic
VQA, hallucinated-future reasoning is \emph{consistently} more discriminative than
observation of the same event at kinematic peak. This is a direct consequence
of JEPA's representational contract: V-JEPA\,2 predicts only the
\emph{latent-space} components that were predictable from context and
\emph{discards} the stochastic pixel-level residual (motion blur, occlusion,
camera shake at impact) that observation encoding must unavoidably ingest. We
formalise this as the \emph{Latent Clarity Hypothesis}: \emph{JEPA-predicted
futures are denoised projections of the physical trajectory and are therefore
more separable on a text-aligned hypersphere than the noisy observed present.}

\subsection{Topological Sinks and Asymmetric Overshoot}
\label{sec:topological-sinks}

The per-category degradation observed under the Cross-Manifold Probe (Table~\ref{tab:per_category}) provides strong empirical verification of the Asymmetric Geometric Overshoot mechanism. When the horizon-specific Anticipation ASP ($\Delta$) is applied to already-sharp Observation latents, the vectors are geometrically ejected from their valid semantic receptive fields, causing categories like Arrest, Arson, Explosion, and Fighting to collapse to absolute $0.0\%$ accuracy. 

Crucially, the preservation or artificial inflation of specific categories (e.g., UCF-Crime Robbery at $66.2\%$, XD-Violence Abuse at $100\%$) does not indicate functional translation. Rather, because the ASP applies a uniform directional shift across the manifold, the ejected vectors pool into a clustered geometric ``dead zone'' within the $\mathcal{S}^{2047}$ hypersphere. Under a closed-set cosine similarity evaluation, the text anchors residing closest to this ejection zone act as topological sinks, absorbing the misclassified mass. The $100\%$ CMP accuracy for XD-Violence Abuse is a stark manifestation of this sink effect acting upon a statistically microscopic sample size ($n=12$). Ultimately, the systematic destruction of the broader 14-class angular separation empirically proves the Anticipation and Observation manifolds occupy geometrically distinct subspaces.

\subsection{Scene bias as the major confound in continuous VAD}
The flatness of Track 1 across \emph{all} horizons, including the
physically-impossible $T{-}5$s, is a warning about naïve continuous VAD:
static CCTV backgrounds (a grocery aisle, a petrol station, an ATM vestibule)
carry strong class priors. The $L_{1}$-surprise z-score gate operationalises a
principled test against this failure.

\subsection{Anticipation vs.\ Autoregressive Generation}
\label{sec:vlm-discussion}

\textbf{Latency \& Alignment.} VLMs decode token-by-token ($>70$\,ms latency) and often refuse to classify violent footage due to RLHF alignment. PULS produces a 2048-d semantic latent state in a single forward pass with no decoder overhead, preserving kinematic severity end-to-end. A successful Y-Decoder (Readout Adapter) confirms the pipeline still supports optional autoregressive decoding when needed or surprise score escalates.

\textbf{Generative Inefficiency \& Incompatibility.} We omit VLM token-generation baselines because the VL-JEPA literature already proves continuous embedding prediction yields a $+108\%$ captioning and $+51\%$ classification advantage over token-space VLMs at equal compute~\cite{vljepa}. Our Y-Decoder thus acts solely as an on-demand semantic readout. Furthermore, PULS structurally rejects pixel-generation evaluations like PBench~\cite{agarwal2025cosmos}. As formalized by the Latent Clarity Hypothesis, forcing pixel-level prediction of aleatoric noise introduces epistemic hallucination and severe latency bottlenecks. By operating strictly within the continuous $\mathcal{S}^{2047}$ manifold, our ASP bypasses both the autoregressive token and pixel reconstruction bottlenecks entirely.

\textbf{Kinematic vs.\ Semantic gating.}
VL-JEPA's semantic variance gating fails on static CCTV, collapsing to scene-bias priors (evidenced by our flat Track~1 accuracy). Our $L_1$-surprise gate thresholds kinematic prediction error \emph{before} the semantic bridge, gating strictly on Newtonian violations --- the necessary adaptation for static-camera predictive anomaly anticipation.

\subsection{Limitations}
\label{sec:limits}
\begin{itemize}[leftmargin=*,nosep]
  \item \textbf{Static-camera bias.} KSD and ASP was distilled on static-CCTV UCF-Crime; XD-Violence ego-motion causes distillation bias (Section~\ref{sec:xd}). Retraining on a mixed corpus is left for future work.
  \item \textbf{Temporal rollout.} The $T{-}2$s result relies on a single pass. Multi-step autoregressive rollout for longer horizons is future work.
  \item \textbf{Encoder provenance.} Upgrading to V-JEPA\,2.1 requires re-distilling the KSD Bridge on the densified latent distribution.
\end{itemize}

\section{Broader Impacts and Deployment}
\label{sec:impacts}
PULS's sub-frame latency enables proactive anomaly anticipation on consumer GPUs. Operating entirely in continuous latent space (no image reconstruction) aligns with biometric privacy regulations (BIPA, GDPR). Furthermore, transmitting 2048-dimensional semantic states rather than raw optical feeds yields orders-of-magnitude reductions in edge-to-cloud bandwidth, enabling high-fidelity monitoring over constrained infrastructure. However, anticipatory kinematic profiling risks flagging lawful behaviour. We mandate human-in-the-loop review for alerts, aided by the Y-Decoder's interpretable reasoning. Discriminatory or political surveillance is strictly unsanctioned.

\section{Conclusion}
\label{sec:conclusion}

In this work, we presented PULS, a framework that shifts the paradigm of video anomaly detection from reactive, pixel-level reconstruction to proactive, latent-space anticipation. By leveraging the self-supervised spatial-temporal representations of Joint Embedding Predictive Architectures (V-JEPA\,2), we demonstrated that a physics-conditioned world model can anticipate critical incident severity and category directly within a Shared Semantic Hypersphere. Through the Kinematic Semantic Distillation (KSD) Bridge, we aligned these high-dimensional video latents with frozen text embeddings, achieving a chunk-level AUROC of $0.8994$ on UCF-Crime and proving robust zero-shot out-of-distribution transfer to XD-Violence ($0.8162$ AUROC). The Anticipatory State Predictor (ASP) further enables proactive classification, yielding a $+14.0$\,pp predictive zero-shot VQA gain and a $+8.9$\,pp lead-time advantage at the critical pre-incident horizon.

Crucially, our experiments validate the \emph{Latent Clarity Hypothesis}: predicting future states within a self-supervised latent space behaves as a denoising projector. Rather than generating high-frequency pixel-level details (which suffer from epistemic blur and sensor noise), the predictive representation discards the stochastic residuals and isolates the underlying causal trajectory. Geometrically, our Cross-Manifold Probe and the resulting Asymmetric Geometric Overshoot confirm that the Observation and Anticipation manifolds occupy distinct, horizon-specific sub-manifolds. The systematic collapse of observation latents under the ASP operation proves that physical anticipation in joint-embedding spaces is a directional vector shift calibrated strictly for the future manifold, preventing topological overlap and semantic interference.

Looking forward, this work opens several avenues for the development of real-world physical intelligence. The integration of $L_1$-surprise gating provides a mathematically principled mechanism to decouple static scene priors from dynamic kinematic anomalies, offering a blueprint for robust, continuous monitoring. By operating entirely in the latent space and transmitting low-bandwidth semantic coordinates instead of raw optical streams, the pipeline is highly suited for edge-device deployment, respecting both bandwidth constraints and user privacy. Future work will focus on retraining the model on a mixed corpus to explicitly mitigate the static-camera distillation bias caused by ego-motion. Furthermore, we aim to extend the single-step rollout to a multi-step autoregressive mechanism, evaluating the statistical drift of the projected latent state over extended horizons (up to $T{-}5.0$\,s) as cumulative epistemic blur degrades accuracy.

\begin{ack}
We gratefully acknowledge the hardware support and compute resources provided by the Vision AI Labs at Sigmind.ai, including the single NVIDIA A40 GPU (48\,GB) footprint used for training and inference. The UCF-Crime and XD-Violence datasets are publicly available for research.
\end{ack}

\bibliographystyle{unsrt}
\bibliography{references}

\clearpage
\begin{center}
{\bf\Large Appendix / Supplementary Material}
\end{center}
\vskip 0.2in

\appendix

\section{Hyperparameters and Training Details}
\label{app:hparams}

\begin{table}[h]
\centering\small
\caption{Full hyperparameter table for KSD Bridge and ASP training.}
\begin{tabular}{lll}
\toprule
\textbf{Parameter} & \textbf{KSD Bridge} & \textbf{ASP} \\
\midrule
Backbone              & LLaMA-3.2-1B (last 8 layers) & Residual MLP (from scratch) \\
Architecture          & Transformer with bidir.\ attn.\ & $\mathrm{Linear}\to\mathrm{GELU}\to\mathrm{Linear}$ + skip \\
Input / hidden / out  & $1024 \to 2048 \to 2048$ & $2048 \to 4096 \to 2048$ \\
Residual init         & N/A & Output layer zero-init \\
Trainable params      & 490M & 16.8M \\
Loss                  & Symmetric InfoNCE ($\tau{=}0.05$) & Symmetric InfoNCE ($\tau{=}0.07$) \\
Teacher signal        & Qwen3-VL text emb.\ (frozen) & KSD($\mathbf{Z}^{\mathrm{obs}}$) (frozen) \\
Noise injection       & $\sigma=0.01$ (SALT) & None \\
Batch size            & 64 & 1024 \\
Optimizer             & AdamW (wd $0.04$) & AdamW (wd $0.01$) \\
Learning rate         & $5\times10^{-5}$ & $3\times10^{-4}$ \\
LR schedule           & Cosine, min $10^{-6}$ & Cosine, min $10^{-6}$ \\
Warmup epochs         & 5 & 5 \\
Total epochs          & 100 & 300 \\
Grad clip             & max-norm $1.0$ & max-norm $1.0$ \\
Training chunks       & 7{,}819 (UCF-Crime) & 7{,}819 (cached pairs) \\
Hardware              & 1$\times$ NVIDIA A40 (48\,GB) & 1$\times$ NVIDIA A40 (48\,GB) \\
Attention             & Bidirectional (4D all-ones mask) & N/A \\
\bottomrule
\end{tabular}
\end{table}

\section{Bidirectional Attention Override}
\label{app:bidir}

HuggingFace \texttt{LlamaModel.forward()} applies a causal (lower-triangular)
attention mask by default. For the KSD Bridge, token position in the sequence
encodes \emph{which part of the clip} (spatial-temporal patches 1--2048) and
\emph{which part of the query} (tokens 2049+), not temporal order. Causal masking
would make every vision token blind to all later tokens and to the query --- a
catastrophic information bottleneck. Our fix:

\begin{verbatim}
# SALT_brigde/services/llama_predictor_service.py:171-184
attention_mask = torch.ones(
    (B, 1, seq_len, seq_len),
    dtype=torch.bool, device=device
)
outputs = llama_model(
    inputs_embeds=x,
    attention_mask=attention_mask,
)
\end{verbatim}

The 4D boolean tensor broadcasts correctly across all 8 Llama layers without
modifying the model weights. This is the only non-standard LLaMA usage in the pipeline.
\end{document}